\documentclass{article}

\usepackage{arxiv}

\usepackage[utf8]{inputenc} 
\usepackage[T1]{fontenc}    
\usepackage{hyperref}       
\usepackage{url}            
\usepackage{booktabs}       
\usepackage{amsfonts}       
\usepackage{nicefrac}       
\usepackage{microtype}      
\usepackage{lipsum}
\usepackage{tablefootnote}
\usepackage{graphicx}
\usepackage{color,soul}
\usepackage{xcolor}
\usepackage{natbib}
\usepackage{cleveref}
\usepackage{subcaption}
\usepackage{makecell}

\title{Enhancing deep neural networks with morphological information}

\author{
    Matej Klemen, Luka Krsnik, Marko Robnik-Šikonja \\
     University of Ljubljana, Faculty of Computer and Information Science \\
     Večna pot 113, Ljubljana, Slovenia \\
    \texttt{mk3141@student.uni-lj.si, krsnik.luka92@gmail.com, marko.robnik@fri.uni-lj.si} \\
}

\begin{document}
\maketitle

\begin{abstract}
Deep learning approaches are superior in natural language processing due to their ability to extract informative features and patterns from languages. The two most successful neural architectures are LSTM and transformers, used in large pretrained language models such as BERT. While cross-lingual approaches are on the rise, most current natural language processing techniques are designed and applied to English, and less-resourced languages are lagging behind. In morphologically rich languages, information is conveyed through morphology, e.g., through affixes modifying stems of words. The existing neural approaches do not explicitly use the information on word morphology. We analyse the effect of adding morphological features to LSTM and BERT models. As a testbed, we use three tasks available in many less-resourced languages: named entity recognition (NER), dependency parsing (DP), and comment filtering (CF). We construct baselines involving LSTM and BERT models, which we adjust by adding additional input in the form of part of speech (POS) tags and universal features. We compare the models across several languages from different language families. Our results suggest that adding morphological features has mixed effects depending on the quality of features and the task.
The features improve the performance of LSTM-based models on the NER and DP tasks, while they do not benefit the performance on the CF task. For BERT-based models, the added morphological features only improve the performance on DP when they are of high quality (i.e. manually checked) while not showing any practical improvement when they are predicted. Even for high-quality features, the improvements are less pronounced in language-specific BERT variants compared to massively multilingual BERT models. As in NER and CF datasets manually checked features are not available, we only experiment with predicted features and find that they do not cause any practical improvement in performance.
\end{abstract}


\section{Introduction}
\label{sec:introduction}
The use of deep learning for processing natural language is becoming a standard, with excellent results in a diverse range of tasks. Two state-of-the-art  neural architectures for text-related modeling are long short-term memory (LSTM) networks~\citep{hochreiter1997-lstm} and transformers~\citep{vaswani2017-transformers}.
LSTMs are recurrent neural networks that sequentially process text one token at a time, building up its internal representation in hidden states of the network.
Due to the recurrent nature of LSTM, which degrades the efficiency of parallel processing, and improvements in performance, models based on the transformer architecture are gradually replacing LSTMs across many tasks. Transformers can process the text in parallel, using self-attention and positional embeddings to model the sequential nature of the text.

A common trend in using transformers is to pre-train them on large monolingual corpora with a general-purpose objective and then fine-tune them with a more specific objective, such as text classification. 
For example, the BERT (Bidirectional Encoder Representations from Transformers) architecture \citep{devlin2019-bert} uses transformers and is pretrained with masked language modelling and order of sentences prediction tasks to build a general language understanding model.
During the fine-tuning for a specific downstream task, additional layers are added to the BERT model, and the model is trained on task-specific data to capture the specific knowledge required to perform the task. 

Most of the research in the natural language processing (NLP) area focuses on English, ignoring the fact that English is specific in terms of the low amount of information expressed through morphology (English is a so-called analytical language).
In our work, we adapt modern deep neural networks, namely LSTM and BERT, for several morphologically rich languages by explicitly including the morphological information. The languages we analyse contain rich information about grammatical relations in the morphology of words instead of in particles or relative positions of words (as is the case in English). For comparison,  we also evaluate our adaptations on English.
Although previous research has shown that the state of the art methods such as BERT already capture some information contained in the morphology~\citep{pires-2019-mtl-bert-probing,edmiston2020systematic, mikhailov-2021-morphcall}, this investigation is commonly done by analysing the internals, for example with probing. Probing studies examine whether a property is encoded inside a model, but not necessarily used. In contrast, we present methods which combine BERT with separately encoded morphological properties: universal part of speech tags (UPOS tags) and universal features (grammatical gender, tense, conjugation, declination, etc.). We evaluate them on three downstream tasks: named entity recognition (NER), dependency parsing (DP), and comment filtering (CF), and observe whether the additional information benefits the models. If it does, the BERT models use the provided additional information, meaning that they do not fully capture it in pretraining. We perform similar experiments on LSTM networks and compare the results for both architectures. 

Besides English, we analyse 10 more languages in NER, 15 in DP, and 5 in CF task. The choice of languages covers different language families but is also determined by the availability of resources and our limited computational resources. 
We describe the data in more detail in \Cref{sec:data}.

Our experiments show that the addition of morphological features has mixed effects depending on the task.
Across the tasks where the added morphological features improve the performance, we show that 1) they benefit the LSTM-based models even if the features are noisy and 2) they benefit the BERT-based models only when the features are of high quality (i.e. human checked), suggesting that BERT models already capture the morphology of the language. We see room for improvement for large pretrained models either in designing pretraining objectives that can capture morphological properties or when high-quality features are available (rare in practice).

The remainder of this paper is structured as follows.
In Section~\ref{sec:related-work}, we present different attempts to use morphological information in the three evaluation tasks and an overview of works studying the linguistic knowledge within neural networks.
In Section~\ref{sec:data}, we describe the used datasets and their properties.
In Section~\ref{sec:methods}, we present the baseline models and models with additional morphological information, whose performance we discuss in Section~\ref{sec:evaluation}.
Finally, we summarize our work and present directions for further research in Section~\ref{sec:conclusion}.

\section{Related work}
\label{sec:related-work}

This section reviews the related work on the use of morphological information within the three evaluation tasks, mainly focusing on neural approaches. We split the review into four parts, one for each of the three evaluation tasks, followed by the works that study the linguistic knowledge contained within neural networks. 

\subsection{Morphological features in NER}
Recent advances in NER are mostly based on deep neural networks.
A common approach to NER is to represent the input text with word embeddings, followed by several neural layers to obtain the named entity label for each word on the output. 
In one of the earlier approaches, \citet{collobert2008-early-dnn-ner} propose an architecture that is jointly trained on six different tasks, including NER, and show that this transfer learning approach generalizes better than networks trained on individual tasks due to learning a joint representation of tasks.
While the authors use Time Delay Neural Network (convolutional) layers \citep{time-delay-neural-networks} to model dependencies in the input, in subsequent works, various recurrent neural networks, such as LSTMs \citep{hochreiter1997-lstm}, are commonly used.
For example, \citet{huang2015-lstm-variants-ner} show that using unidirectional and bidirectional LSTM layers for NER, results in comparable or better performance than approaches using convolutional layers.

In addition to word embeddings, these systems often use hand-crafted features, such as character-based features or affixes.
The work of \citet{dos-santos-2015-char-embeddings-ner} outlines the inconvenience of constructing such features and proposes their automatic extraction using character embeddings and a convolutional layer. Authors combine the character-level features with word-level features to obtain competitive or improved results on Spanish and Portuguese NER.
Multiple authors \citep{kuru2016-charner, lample-2016-neuralner, yang2016-multitask-ner} confirm the effectiveness of character embeddings and show that recurrent layers can be used to process them instead of convolutional layers.

While sequence modelling on the character level can already encode morphological information, several authors show that the performance of neural networks on the NER task can be improved by including additional information about the morphological properties of the text.
\citet{strakova2016-ner-pos} present a Czech NER system that surpasses the previous best system using only form, lemma and POS tag embeddings. Similarly, \citet{gungor2017-turkish-ner} show that using morphological embeddings in addition to character and word embeddings improves performance on Turkish and Czech NER, while \citet{simeonova2019-bulgarian-ner} show that including additional morphological and POS features improves performance on Bulgarian NER. \citet{gungor2017-turkish-ner-2} extend the study of \citet{gungor2017-turkish-ner} to three additional languages: Hungarian, Finnish and Spanish.

The influence of POS tags and morphological information on the NER performance of BERT models is less studied.
\citet{nguyen-2021-phonlp} present a multi-task learning model which is jointly trained for POS tagging, NER, and DP. Their multi-task learning approach can be seen as an implicit injection of additional POS tags into BERT. The system trained on multiple tasks outperforms the respective single-task baselines, indicating that the additional information is beneficial.
More explicitly, \citet{mohseni-2019-morphobert} use morphological analysis as a preprocessing step to split the words into lemmas and affixes before passing them into a BERT model. Their system achieved first place in the NER shared task organized as part of the Workshop on NLP solutions for under-resourced languages 2019 \citep{taghizadeh-2019-nsurl-task7}. However, they do not ablate the morphological analysis component, so it is unclear exactly how helpful it is in terms of the NER performance.

Our work extends the literature that studies the influence of POS tags and morphological information on the NER performance. For LSTM models, some works already explore this impact, though they are typically limited to one or a few languages. At the same time, we perform a study on a larger pool of languages from different language families. 
For BERT, we are not aware of any previous work that studies the influence of explicitly including POS tags or morphological information on the performance of the NER task.
The existing work either adds the information implicitly \citep{nguyen-2021-phonlp} or does not show the extent to which the additional information is useful \citep{mohseni-2019-morphobert}. In addition, existing analyses are limited to a single language. In contrast, our work explicitly adds POS tags or morphological information and shows how it affects the downstream (NER) performance on a larger pool of languages.

\subsection{Morphological features in dependency parsing}
Similarly to NER, recent progress in DP is dominated by neural approaches. Existing approaches introduce neural components into either a transition-based \citep{yamada-matsumoto-2003-transition-based-origin-1,nivre-2003-transition-based-origin-2} or a graph-based parser \citep{mcdonald-2005-graph-based-origin}. Some works do not fall into either category, e.g., they treat DP as a sequence-to-sequence task \citep{li-2018-dep-parsing-seq2seq}. 
The two categories differ in how dependency trees are produced from the output of prediction models. 
In the transition-based approach, a model is trained to predict a sequence of parsing actions that produce a valid dependency tree. In contrast, in the graph-based approach, a model is used to score candidate dependency trees via the sum of scores of their substructures (e.g., arcs).

One of the earlier successful approaches to neural DP was presented by  \citet{chen-manning-2014-early-nn-transition}, who replaced the commonly used sparse features with dense embeddings of words, POS tags and arc labels, in combination with the transition-based parser. This approach improved both the accuracy and parse speed.
\citet{pei-2015-early-nn-graph} introduced a similar approach to graph-based parsers.
Later approaches improve upon the earlier methods by automatically extracting more information that guides the parsing, e.g., researchers use LSTM networks to inject context into local embeddings \citep{kiperwasser-goldberg-2016-lstm-parser}, apply contextual word embeddings \citep{kulmizev2019-contextual-embeddings-parser}, or train graph neural networks \citep{ji-2019-gnn-parser}.

The use of morphological features in DP, especially for morphologically rich languages, is common and predates neural approaches. 
For example, \citet{marton-2010-arabic-parser} study the contribution of morphological features for Arabic, \citet{seeker-2011-german-parser} for German, \citet{kapovciute-2013-lithuanian-parser} for Lithuanian, \citet{khallash-2013-persian-parser} for Persian, etc.
The majority of such works report improved results after adding morphological information.
As our focus is on neural approaches, we mostly omit pre-neural approaches. Still, we note that the research area is extensive and has also been the topic of workshops such as the Workshop on statistical parsing of morphologically rich languages (\cite{spmrl-2010}).
The largely positive results have motivated authors to continue adding morphological information also to neural systems, which already automatically learn features and may pick up this information. 
For example, \citet{chen-manning-2014-early-nn-transition} note that POS tag embeddings contribute to the strong performance of their neural system on English and Chinese, and \citet{ozates-2018-morphological-embeddings} note the usefulness of morphological embeddings for multiple agglutinative languages. 
\citet{dozat-2017-stanfordconll17} report a similar trend in their work submitted to the CoNLL 2017 shared task \citep{conll17sharedtask}. They emphasize that POS tags are helpful, but only if produced using a sufficiently accurate POS tagger.

Similarly as in NER, character embeddings improve the accuracy of LSTM-based dependency parsers. We use the term ``LSTM-based'' to refer to models that include an LSTM neural network (as opposed to a more recent transformer neural network \citep{vaswani2017-transformers}).
Several authors \citep{ballesteros-2015-char-embs-dp, dozat-2017-stanfordconll17, delhoneux-2017-uppsalaconll17} report that the use of character-based embeddings results in an improvement in the DP performance and can act as an approximate replacement for additional morphological information.
Whether the embeddings present an approximate or complete replacement of morphological information is not entirely certain: \citet{vania-2018-char-embs-morphology} show that models using character embeddings can still benefit from additional inclusion of morphological features. In contrast, \citet{anderson-2020-fraility-upos} report that the addition of POS tag embeddings does not further help a parser using character embeddings unless the POS tags are of a practically unrealistic quality.

Multi-task learning approaches, where a model is jointly trained for dependency parsing and another task (such as POS- or morphological tagging), are also a common way to inject additional information into dependency parsers \citep{straka-2018-udpipe, lim-2018-sexbist, nguyen-verspoor-2018-mtl-upos-dep}. 
Such approaches are popular in BERT-based parsers \citep{kondratyuk-straka-2019-multilingual-multitask-bert-parser, zhou2020-is-pos-necessary, lim2020-multiview-multitask-pos-dp, grunewald-2021-occam-dp} and seem to be more common than the alternative approach of using additional inputs in a single task DP system. In our work, we use the alternative approach and explicitly include POS tags and morphological features in the form of their additional embeddings. 
We test the effect of the additional information on a sizable pool of languages from diverse language families. We perform several experiments to provide additional insight, testing the same effect with longer training, noisy information, and language-specific BERT models.

\subsection{Morphological features in comment filtering}
The literature for the CF task covers multiple related tasks, such as hate speech, offensive speech, political trolling, detecting commercialism, etc. Recent approaches involve variants of deep neural networks, though standard machine learning approaches are still popular, as shown in the survey of \citet{fortuna-2018-hate-speech-survey}. These approaches typically use features such as character n-grams, word n-grams, and sentiment of the sequence. Two examples are the works of \citet{malmasi-2017-hate-speech-classic-ml-1}, who classify hate speech in English tweets, and \citet{van-hee-2015-dutch-cyberbullying}, who classify different levels of cyber-bullying in Dutch posts on ask.fm social site.
\citet{scheffler-2018-hate-speech-embeddings} combine word embeddings with the features mentioned above to classify German tweets. They observe that combining both n-gram features and word embeddings brings only a small improvement over only using one of them. The effectiveness of using features describing syntactic dependencies for toxic comments classification on English Wikipedia comments is shown by \citet{shtovba-2019-hate-speech-dependencies}.

Neural architectures used include convolutional neural networks \citep{georgakopoulos-2018-hate-speech-cnn} and LSTM networks \citep{gao-2017-hate-speech-lstm,miok2019prediction}, typically improving the performance over standard machine learning approaches.
The CF topic has also been the focus of shared tasks on identification and categorization of offensive language \citep{zampieri-2019-hate-speech-semeval6} and multilingual offensive language identification \citep{offenseval2020}.
The reports of these tasks show the prevalence and general success of large pretrained contextual models such as BERT, though, surprisingly, the best performing model for the subtask B of SemEval-2019 Task 6 was rule-based \citep{han-etal-2019-jhan014}. 

\subsection{Linguistic knowledge combined with neural networks}
Large pretrained models such as BERT show superior performance across many tasks. Due to a lack of theoretical understanding of this success, many authors study how and to what extent BERT models can capture various information, including different linguistic properties. An overview of recent studies in this area, sometimes referred to as BERTology, is compiled by \citet{rogers2020primer}.
Two common approaches to study BERT are i) add additional properties to BERT models and observe the difference in performance on downstream tasks, ii) a technique called probing~\citep{conneau-etal-2018-probing}, where the BERT model is trained (fine-tuned) to predict a studied property. As we have noted examples of i) in previous sections, we focus on the probing attempts here.

For example, \citet{jawahar-etal-2019-bert-language-structure} investigate what type of information is learned in different layers of the BERT English model and find that it captures surface features in lower layers, syntactic features in middle layers, and semantic features in higher layers.
Similarly,~\cite{lin2019-bert-linguistic} find that BERT encodes positional information about tokens in lower layers and then builds increasingly abstract hierarchical features in higher layers.
~\cite{tenney-etal-2019-bert-rediscovers-pipeline} use probing to quantify where different types of linguistic properties are stored inside BERT's architecture and suggest that BERT implicitly learns the steps performed in classical (non-end-to-end) NLP pipeline. However, \cite{elazar2020-amnesic-probing}  point out possible flaws in the probing technique, suggesting amnesic probing as an alternative.
They arrive at slightly different conclusions about BERT layer importance; for example, they show that the POS information greatly affects the predictive performance in upper layers.

Probing studies for morphological properties were conducted by \citet{edmiston2020systematic} and \citet{mikhailov-2021-morphcall}. Concretely, they train a classifier to predict morphological features based on hidden layers of BERT. Based on the achieved high performance, \citet{edmiston2020systematic} argues that monolingual BERT models capture significant amounts of morphological information and partition their embedding space into linearly-separable regions, correlated with morphological properties.
\citet{mikhailov-2021-morphcall} extend this work to multiple languages, performing probing studies on multilingual BERT models.

\section{Data}
\label{sec:data}
In this section, we describe the datasets used in our experiments separately for each of the three tasks: NER, DP, and CF.

\subsection{Named entity recognition}
In the NER experiments, we use datasets in 11 languages from different language families:  Arabic, Chinese, Croatian, English, Estonian, Finnish, Korean, Latvian, Russian, Slovene and Swedish.
The number of sentences and tags present in the datasets is shown in Table~\ref{tab:ner-datasets}.
The label sets used in datasets for different languages vary, meaning that some contain more fine-grained labels than others.
To make results across different languages consistent, we use IOB encoded labels present in all datasets: location (B/I-LOC), organization (B/I-ORG), person (B/I-PER), and ``no entity'' (O). We convert all other labels to the ``no entity'' label (O).

\begin{table}[htb]
    \centering
	\caption{The collected datasets for NER task and their properties: the number of sentences and tagged words. We display the results for the languages using their ISO 639-2 three letter code, provided in the ``Code'' column.}
	\label{tab:ner-datasets}
	\begin{tabular}{lllrrcl}
		\toprule
		Language	& Code & Dataset & Sentences & Tags \\ 
		\midrule
		Arabic	& ARA & ANERCorp \citep{benajiba-2007-anercorp} & $5005$ & $14876$ \\
		Chinese	& ZHO & MSRA \citep{levow-2006-msra} & $48441$ & $261940$ \\
		Croatian & HRV & hr500k \citep{ljubesic-2018-hr500k} & $24794$ & $28902$ \\
		English & ENG & CoNLL-2003 NER \scriptsize{\citep{tjong-kim-sang-de-meulder-2003-conll2003NER}} & $20744$ & $43979$ \\
		Estonian & EST & Estonian NER corpus \citep{tkachenko-2013-estonianner} & $14287$ & $20965$ \\
		Finnish & FIN & FiNER data \citep{ruokolainen-2019-finer} & $14484$ & $16833$ \\
		Korean & KOR & KMOU NER & $3659$ & $6635$ \\
		Latvian & LAV & LV Tagger train data \citep{paikens-2012-lvner} & $9903$ & $11599$ \\
		Russian & RUS & factRuEval-2016  \citep{starostin-2016-FactRuEval2E} & $4907$ & $9666$ \\
		Slovene\tablefootnote{The Slovene ssj500k originally contains more sentences, but only $9489$ are annotated with named entities.} & SLV & ssj500k \citep{krek-2019-ssj500k} & $9489$ & $9440$ \\
		Swedish	& SWE & Swedish NER & $9369$ & $7292$ \\
		\bottomrule
	\end{tabular}
\end{table}

\subsection{Dependency parsing}
To test morphological neural networks on the DP task, we use datasets in 16 languages from different language families: Arabic, Chinese, Croatian, English, Estonian, Finnish, Hebrew, Hungarian, Korean, Latvian, Lithuanian, Persian, Russian, Slovene,  Swedish, and Turkish. 
We use the datasets from the Universal Dependencies \citep{universal-dependencies}, which contain a collection of texts annotated with UPOS tags, XPOS tags (fine-grained POS), universal features, and syntactic dependencies. We provide the summary of the datasets in Table~\ref{tab:depparse-datasets}. 
The splits we use are predefined by the authors of the datasets. While most of the annotations are manually verified, the universal features in the Chinese and English datasets and the UPOS tags and universal features in the Turkish dataset are only partially manually verified. For Korean, the dataset does not contain universal feature annotations.

\begin{table}[htb]
    \centering
	\caption{Dependency parsing datasets and their properties: the treebank, number of tokens, number of sentences, and the information about the size of splits. We display the dataset information using the language ISO 639-2 three-letter code provided in the ``Code'' column.
	}
	\label{tab:depparse-datasets}
	\begin{tabular}{lllrrrrr}
	\toprule
	Language & Code & Treebank & Tokens & Sentences & Train & Validation & Test \\
	\midrule
	Arabic & ARA & PADT & $282384$ & $7664$ & $6075$ & $909$ & $680$ \\
	Chinese & ZHO & GSD & $123291$ & $4997$ & $3997$ & $500$ & $500$ \\
	Croatian & HRV & SET & $199409$ & $9010$ & $6914$ & $960$ & $1136$ \\
	English & ENG & EWT & $254855$ & $16622$ & $12543$ & $2002$ & $2077$ \\
	Estonian & EST & EDT & $438171$ & $30972$ & $24633$ & $3125$ & $3214$ \\
	Finnish & FIN & TDT & $202697$ & $15135$ & $12216$ & $1364$ & $1555$ \\
	Hebrew & HEB & HTB & $161411$ & $6216$ & $5241$ & $484$ & $491$ \\
	Hungarian & HUN & Szeged & $42032$ & $1800$ & $910$ & $441$ & $449$ \\
	Korean & KOR & Kaist & $350090$ & $27363$ & $23010$ & $2066$ & $2287$ \\
	Latvian & LAV & LVTB & $220536$ & $13643$ & $10156$ & $1664$ & $1823$ \\
	Lithuanian & LIT & ALKSNIS & $70051$ & $3642$ & $2341$ & $617$ & $684$ \\
	Persian & FAS & PerDT & $501776$ & $29107$ & $26196$ & $1456$ & $1455$ \\
	Russian & RUS & GSD & $98000$ & $5030$ & $3850$ & $579$ & $601$ \\
	Slovene & SLV & SSJ & $140670$ & $8000$ & $6478$ & $734$ & $788$ \\
	Swedish & SWE & Talbanken & $96858$ & $6026$ & $4303$ & $504$ & $1219$ \\
	Turkish & TUR & BOUN & $122383$ & $9761$ & $7803$ & $979$ & $979$ \\
	\bottomrule
	\end{tabular}
\end{table}

\subsection{Comment filtering}
While comparable datasets exist across different languages for the NER and DP task, no such standard datasets exist for the CF task. For that reason, in our experiments on CF, we select languages for which adequate datasets exist, i.e. large, of sufficient quality, and reasonably balanced across classes. We provide a summary of the used datasets in Table \ref{tab:comment-filtering-datasets}. 

For English experiments, we use a subset of toxic comments from Wikipedia's talk page edits used in Jigsaw's toxic comment classification challenge \citep{jigsaw-toxic-comments-annotation}.
The comments are annotated with six possible labels: toxic, severe toxic, obscene language, threats, insults, and identity hate (making a total of six binary target variables). We extracted comments from four categories: toxic, severe toxic, threats, and identity hate, a total of $21,541$ instances. We randomly chose the same amount of comments that do not fall in any of the mentioned categories, obtaining the final dataset of $43,082$ instances, using $60\%$ randomly selected examples as the training, $20\%$ as the validation, and $20\%$ as the test set.

For Korean experiments, we use a dataset of comments from a Korean news platform \citep{moon-2020-korean-hate}, annotated as offensive, hateful or clean. We group the offensive and hateful examples to produce a binary classification task.
As the test set labels are private, we instead use the predefined validation set as the test set and set aside $20\%$ of the training set as the new validation set.

For Slovene experiments, we use the IMSyPP-sl dataset \citep{imsypp-1}, containing tweets annotated for fine-grained hate speech. The tweets are annotated with four possible labels: appropriate (i.e. not offensive), inappropriate, offensive, or violent. Each tweet is annotated twice, and we only keep a subset for which both labels agree. To produce a binary classification task, we group the tweets labelled as inappropriate, offensive, or violent into a single category.
We use the predefined split into a training and test set, and additionally, remove $20\%$ of examples from the training set for use in the validation set.

For Arabic, Greek, and Turkish experiments, we use datasets from the OffensEval 2020 shared task on multilingual offensive language identification \citep{offenseval2020}. The datasets are composed of tweets annotated for offensive language: a tweet is either deemed offensive or not offensive.
We use the predefined splits into training and test sets provided by the authors. We randomly remove $20\%$ of the examples from the original training sets to create the validation sets.
As the Turkish training set proved to be too heavily imbalanced, we decided to randomly remove half of the unoffensive examples from it before creating the validation set.

\begin{table}[htb]
    \centering
	\caption{Comment filtering datasets and their properties: number of examples, size of the split and class distribution inside subsets. 
	}
	\label{tab:comment-filtering-datasets}
	\resizebox{\linewidth}{!}{
	\begin{tabular}{llrrrrr}
	\toprule
	Language & Dataset & Examples & Train & Validation & Test \\
	\midrule
	Arabic & OffensEval 2020 \citep{offenseval2020}& $9959$ & $6370$ & $1597$ & $1992$ \\
	English  & Jigsaw toxic comments \citep{jigsaw-toxic-comments-annotation}& $43082$ & $25848$ & $8616$ & $8618$ \\
	Greek  & OffensEval 2020 \citep{offenseval2020}& $10287$ & $6994$ & $1749$ & $1544$ \\
	Korean  & Korean hate speech dataset \citep{moon-2020-korean-hate}& $8367$ & $6316$ & $1580$ & $471$ \\
	Slovene  & IMSyPP-sl\citep{imsypp-1} & $47538$ & $31676$ & $7919$ & $7943$ \\
	Turkish  & OffensEval 2020 \citep{offenseval2020}& $22470$ & $15154$ & $3789$ & $3527$ \\
	\bottomrule
	\end{tabular}
	}
\end{table}

\section{Neural networks with morphological features}
\label{sec:methods}
This section describes the architectures of neural networks used in our experiments. Their common property is that we enhance standard word embeddings based inputs with embeddings of morphological features. We work with recent successful neural network architectures, LSTMs and transformers, i.e. BERT models. A detailed description of architectures is available in the following subsections, separately for each evaluation task. We describe the baseline architecture and the enhanced one for each task and architecture.

\subsection{Named entity recognition models}
\label{sec:ner-models}

\begin{figure}[htb]
    \centering
    \includegraphics[width=0.7\textwidth]{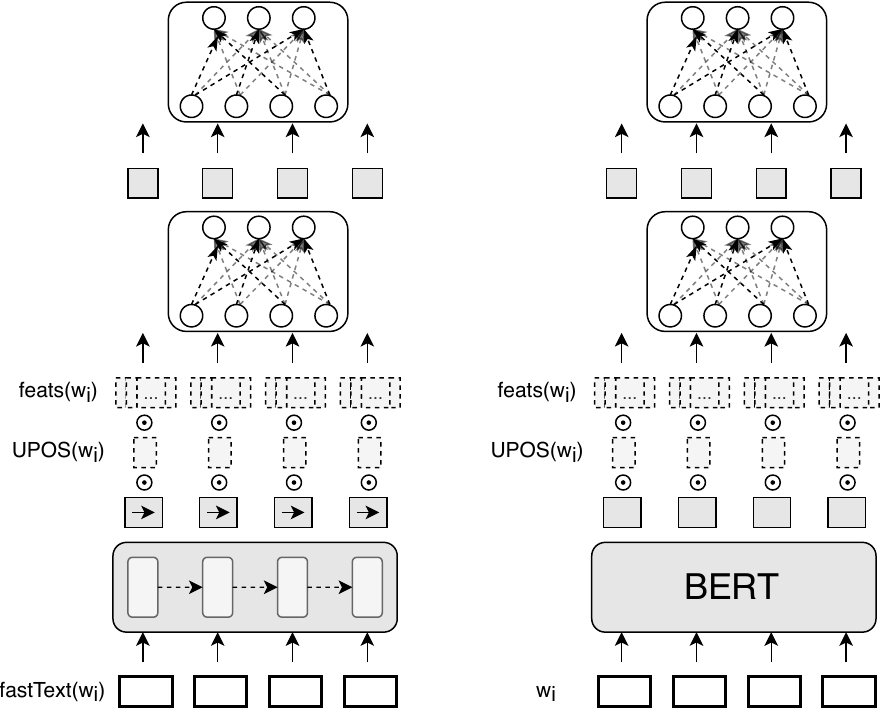}
    \vspace*{5mm}
    \caption{The baseline LSTM-based (left) and BERT-based (right) models for the NER task, along with our modifications with morphological information. The dotted border of POS vectors and morphological feature vectors (feats) marks that their use is optional and varies across experiments. The $\odot$ symbol between layers represents the concatenation operation. The $w_i$ symbol stands for token $i$; in case of LSTM, tokens enter the model sequentially, and we show the unrolled network, while BERT processes all tokens simultaneously.}
    \label{fig:ner_models}
\end{figure}

In the NER task, we use two baseline neural networks (LSTM and BERT) and the same two models with additional morphological information: POS tag embeddings and universal feature embeddings. The baseline models and their enhancements are displayed in Figure~\ref{fig:ner_models}.

The first baseline model (left-hand side of \Cref{fig:ner_models}) is a unidirectional (left-to-right) LSTM model, which takes as an input a sequence of tokens, embedded using $300$-dimensional fastText embeddings \citep{bojanowski2017enriching}. These embeddings are particularly suitable for morphologically rich languages as they work with subword inputs\footnote{The precomputed embeddings are available at \href{https://fasttext.cc/docs/en/crawl-vectors.html}{https://fasttext.cc/docs/en/crawl-vectors.html}.}.
For each input token, its LSTM hidden state is extracted and passed through the linear layer to compute its tag probabilities.

The second baseline model (right-hand side of \Cref{fig:ner_models}) is the cased multilingual BERT base model (bert-base-multilingual-cased). In our experiments, we follow the sequence tagging approach suggested by the authors of BERT~\citep{devlin2019-bert}. The input sequence is prepended with a special token [CLS] and passed through the BERT model. The output of the last BERT hidden layer is passed through the linear layer to obtain the predictions for NER tags. 

Both baseline models (LSTM and BERT) are enhanced with the same morphological information: POS tag embeddings and universal feature embeddings for each input token. We embedded the POS tags using $5$-dimensional embeddings. For each of the $23$ universal features used (we omitted the \textit{Typo} feature, as the version of the used tagger did not annotate this feature), we independently constructed $3$-dimensional embeddings, meaning that we obtained a $69$-dimensional universal embedding. 
We embedded the features independently due to a large number of their combinations and treated them equally in the DP and CF experiments. We selected the size of the embeddings based on the results of preliminary experiments on Slovene and Estonian languages. We automatically obtained the POS tags and morphological features using the Stanza system \citep{qi2020stanza}.
In the enhanced architectures, we included another linear layer before the final linear classification layer to model possible interactions.

\subsection{Dependency parsing models}
\label{sec:depparse-models}

\begin{figure}
    \centering
    \includegraphics[width=0.7\textwidth]{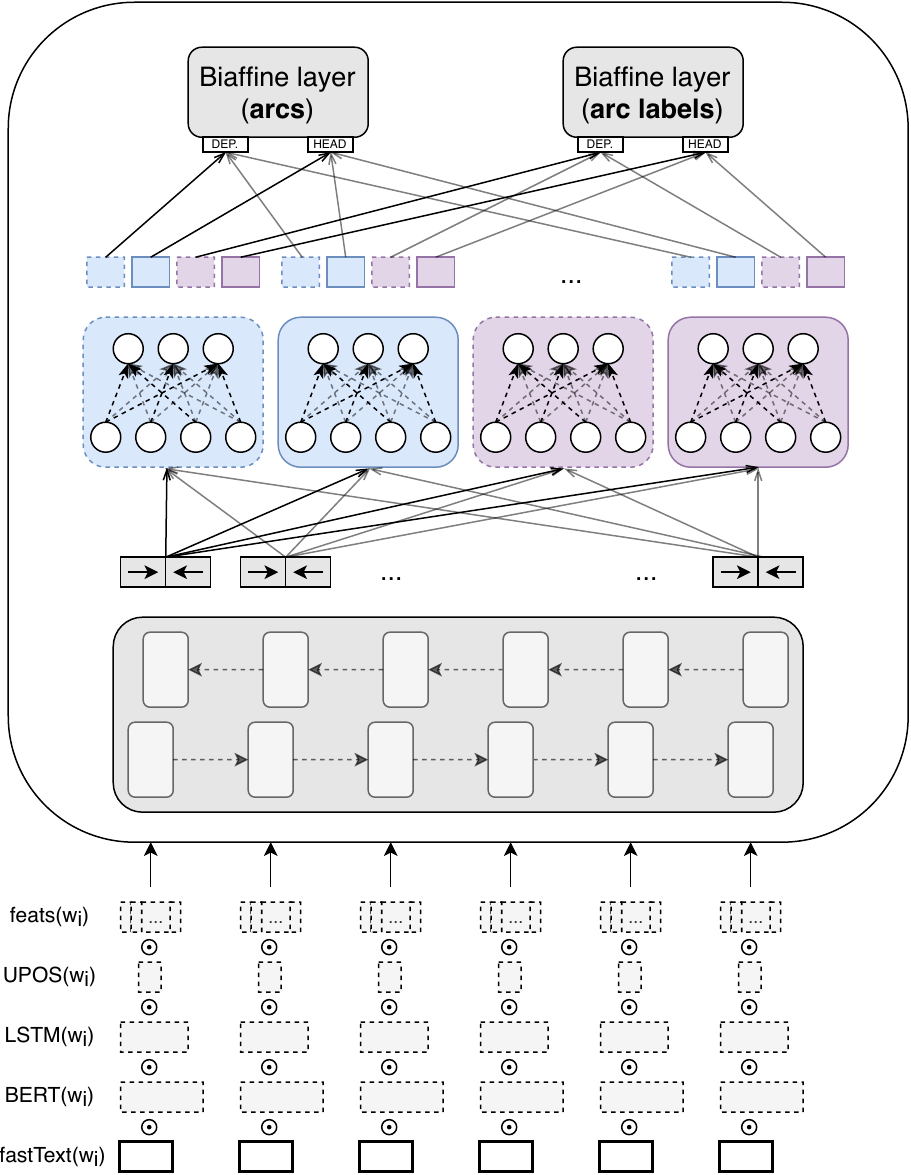}
    \vspace*{5mm}
    \caption{The deep biaffine graph-based dependency parser along with our enhancements at the input level. 
    The dotted border of input embedding vectors, POS vectors, and morphological features (feats) is optional and varies across experiments. The $\odot$ symbol between layers represents the concatenation operation. The $w_i$ symbol stands for token $i$; tokens enter the LSTM model sequentially, and we show the unrolled network.}
    \label{fig:dependency_parsing}
\end{figure}

As the baseline model in the DP task, we use the deep biaffine graph-based dependency parser \citep{dozat2016deep}. The enhancements with the morphological information are at the input level. The baseline model and its enhancements are shown in Figure~\ref{fig:dependency_parsing}.

The baseline parser combines a multi-layer bidirectional LSTM network with a biaffine attention mechanism to jointly optimize prediction of arcs and arc labels.
We leave the majority of baseline architectural hyperparameters at values described in the original paper ($3$-layer bidirectional LSTM with $100$-dimensional input word embeddings and the hidden state size of $400$).

In our experiments, we concatenate the non-contextual word embeddings with various types of additional information.
The first additional input is contextual word embeddings, which we obtain either by using the hidden states of an additional single-layer unidirectional LSTM or by using a learned linear combination of all hidden states of an uncased multilingual BERT base model (bert-base-multilingual-uncased). To check whether the results depend on using cased or uncased BERT model, we rerun experiments for a small sample of languages with the cased version, using identical hyperparameters, and present the results in Appendix \ref{appendix:cased-mtl-bert}. The conclusions drawn from both types are similar; the improvements of enhanced models are statistically insignificant for only one out of the eight sampled languages when using a cased BERT model. 

Although the LSTM layers are already present in the baseline parser, we include an additional LSTM layer at the input level to explicitly encode the context, keeping the experimental settings similar across our three evaluation tasks, i.e. we have one setting with added LSTM and one setting with added BERT.
The second additional input is universal POS embeddings (UPOS), and the third is universal feature embeddings (feats). These embeddings are concatenated separately for each token of the sentences. The size of the additional LSTM layer, POS tag embeddings and universal feature embedding are treated as tunable hyperparameters.
As the baseline input embeddings, we use pre-trained $100$-dimensional fastText embeddings, which we obtain by reducing the dimensionality of publicly available $300$-dimensional vectors with fastText's built-in dimensionality reduction tool. 

In DP experiments, we use POS tags and morphological features of two origins. The first source is human annotations provided in the used datasets. The second source of morphological information is predictions of Stanza models \citep{qi2020stanza}. These two origins are used to assess the quality of morphological information; namely, we check if manual human annotations provide any benefit compared to automatically determined POS tags and features.

\subsection{Comment filtering models}
\label{sec:comment-filtering-models}
\begin{figure}
    \centering
    \includegraphics[width=0.7\textwidth]{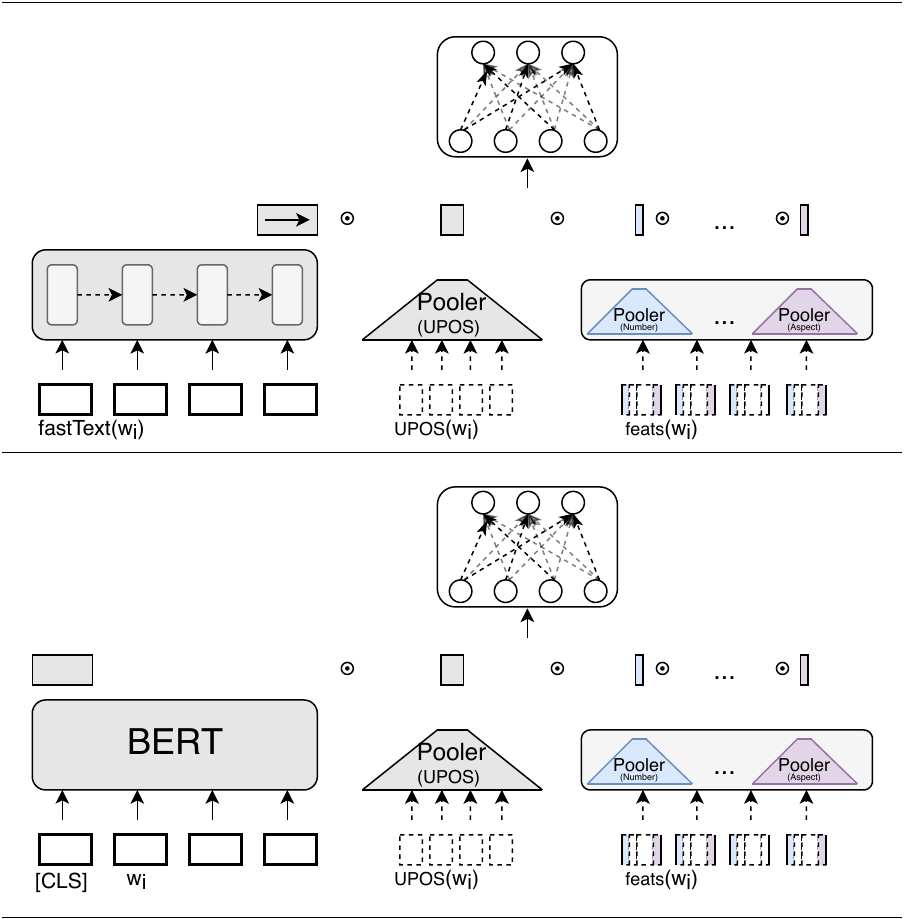}
    \vspace*{5mm}
    \caption{The baseline LSTM (top) and BERT (bottom) models for the CF task, along with our modifications with morphological information. The dotted border of UPOS vectors and morphological features (feats)  marks that their use is optional and varies across experiments. The $\odot$ symbol between layers represents the concatenation operation. The $w_i$ symbol stands for token $i$; in the case of LSTM, tokens enter the model sequentially, and we show the unrolled network, while BERT processes all tokens simultaneously.}
    \label{fig:comment_filtering}
\end{figure}

We add additional morphological information to standard LSTM and BERT models in the CF evaluation task. The baseline and enhanced models for this task are similar to those in the NER evaluation, though we operate at the sequence level here instead of the token level in the NER task. The architecture of models is shown in Figure~\ref{fig:comment_filtering}. 
As baselines, we take a single layer unidirectional LSTM network (the top part of Figure~\ref{fig:comment_filtering}) and the multilingual base uncased BERT model (bert-base-multilingual-uncased; the bottom part of Figure~\ref{fig:comment_filtering}). The difference in the used BERT dialect (\emph{uncased} as opposed to the \emph{cased} in the NER task) is due to better performance detected in preliminary experiments. 

In the LSTM baseline model, the words of the input sequence are embedded using pre-trained $300$-dimensional fastText embeddings. As the representation of the whole sequence, we take the output of the last hidden state, which then passes through the linear layer to obtain the prediction scores.
In the BERT baseline model, we take the sequence classification approach suggested by the authors of BERT. The input sequence is prepended with the special [CLS] token and passed through BERT. The sequence representation corresponds to the output of the last BERT hidden layer for the [CLS] token, which is passed through a linear layer to obtain the prediction scores. 

We augment the baseline models with POS tags and universal feature embeddings, as in other tasks. We obtain the tags for each token separately using the Stanza system \citep{qi2020stanza}. 
We combine the obtained embeddings using three different pooling mechanisms: mean, weighted combination, or LSTM pooling.
Given the POS tag or universal feature embeddings, the mean pooling outputs the mean of all token embeddings. The weighted pooling outputs the weighted combination of token embeddings, and the LSTM pooling outputs the last hidden state obtained by passing the sequence of embeddings through the LSTM network. Both the embedding sizes and the type of pooling are treated as tunable hyperparameters. 
The coefficients of the weighted combination are learned by projecting the sequence embeddings into a sequence of independent dimension values, which are normalized with the softmax. This compresses the embedding sequence, establishes its fixed length, and allows different morphological properties to have a different impact on the sequence representation. This approach tests the contextual encoding of morphological properties. For example, we might learn that the adjectives are assigned a higher weight than other POS tags due to their higher emotional contents that often indicates insults.

\section{Evaluation}
\label{sec:evaluation}
In this section, we first present the evaluation scenario for the three evaluation tasks, followed by the results presented separately for each of the tasks. We end the section with additional experiments performed on the DP task. 
In the additional experiments, we investigate the effect of additional morphological features in three situations where we tweak one aspect of the training procedure at a time: 1) we increase the maximum training time of our models by additional $5$ epochs, 2) we replace the human-annotated features with the ones automatically predicted by machine learning models, and 3) we replace the embeddings from general multilingual BERT with those from more specific multilingual and monolingual BERT models.

\subsection{Experimental settings}
The experimental settings differ between the three evaluation tasks, so we describe them separately for each task, starting with the NER task and then the DP and CF tasks.
One aspect that is common across all three tasks is that we skip the experiments involving universal features on Korean, as the features are not available for any of the bigger Korean Universal Dependencies corpora; therefore, the Stanza models cannot be used to predict them either.

\subsubsection{Experimental settings for NER}
 For NER, we train each BERT model for $10$ epochs and each LSTM model for $50$ epochs. These parameters were determined during preliminary testing on the Slovene dataset. The selected numbers of epochs are chosen to balance the performance and training times of the models. All NER models are evaluated using $10$-fold cross-validation.

We assess the performance of the models with the $F_1$ score, which is a harmonic mean of precision and recall measures. This measure is typically used in NER in a way that precision and recall are calculated separately for each of the three entity classes (location, organization, and person, but not ``no entity''). We compute the weighted average over the class scores for each metric, using the frequencies of class values in the datasets as the weights. Ignoring the ``no entity'' (O) label is a standard approach in the NER evaluation and disregards words that are not annotated with any of the named entity tags, i.e. the assessment focuses on the named entities.

\subsubsection{Experimental settings for dependency parsing}
We train each DP model for a maximum of $10$ epochs, using an early stopping tolerance of $5$ epochs. All models are evaluated using predefined splits into training, validation and testing sets, determined by the respective treebank authors and maintainers (see \Cref{tab:depparse-datasets}).
The final models, evaluated on the test set, are selected based on the maximum mean of unlabeled and labelled attachment scores (UAS and LAS) on the validation set. The UAS and LAS are standard accuracy metrics in DP. The UAS score is defined as the proportion of tokens assigned the correct syntactic head. In contrast, the LAS score is the proportion of tokens that are assigned the correct syntactic head as well as the dependency label \citep{speech-and-language-processing}. As both scores are strongly correlated in most models, we only report the LAS scores to make the presentation clearer.

\subsubsection{Experimental settings for comment filtering}
In our CF experiments, we train BERT models for a maximum of $10$ epochs and LSTM models for a maximum of $50$ epochs, using early stopping with the tolerance of 5 epochs.

We evaluate the models using the macro $F_1$ score, computed as the unweighted mean of the $F_1$ scores for each label. We use fixed training, validation and test sets, described in Table \ref{tab:comment-filtering-datasets}. As we have observed noticeable variance in the preliminary experiments, we report the mean metrics over five runs for each setting, along with the standard deviation.

\subsection{Experimental results}
In this section, we compare the results of baseline models with their enhancements using additional morphological information. We split the presentation into three parts, according to the evaluation task: NER, DP, and CF.

\subsubsection{Results for the NER task}
\label{sec:ner-results}
For the NER evaluation task, we present results of the baseline NER models and models enhanced with the POS tags and universal features, as introduced in \Cref{sec:ner-models}. Table \ref{tab:ner-results} shows the results for LSTM and BERT models for $11$ languages. 
We compute the statistical significance of the differences between the baseline LSTM and BERT models and their best-performing counterparts with morphological additions. We use the Wilcoxon signed-rank test \citep{wilcoxon1970critical} and underline the statistically significant differences at $p=0.01$ level.

\setlength{\tabcolsep}{3pt}
\begin{table}[htb]
	\centering
	\caption{$F_1$ scores of different models on the NER task in 11 languages. The left part of the table shows the results for the LSTM models and the right part for the BERT models. The best scores for each language and neural architecture are marked with the bold typeface. Best scores for which the difference to the respective baseline is statistically significant are underlined.}
	\label{tab:ner-results}
	\begin{subtable}{0.47\textwidth}
    \centering
    \caption{LSTM.}
    \begin{tabular}{l @{\hskip 20pt} cccc}
    \toprule
         lang. &
         \makecell{+LSTM\\\vphantom{}\\\vphantom{}} & 
         \makecell{+LSTM\\+UPOS\\\vphantom{}} & 
         \makecell{+LSTM\\+UPOS\\+feats} & 
         \makecell{+LSTM\\+feats\\\vphantom{}} \\
    \midrule
    ARA & 
    $0.690$ & 
    $0.690$ &
    $0.689$ & 
    $\textbf{0.691}$ \\
    ENG & 
    $\textbf{0.890}$ & 
    $\textbf{0.890}$ & 
    $0.884$ & 
    $0.886$ \\
    EST & 
    $0.769$ & 
    $\underline{\textbf{0.789}}$ & 
    $0.786$ & 
    $0.777$ \\
    FIN & 
    $0.814$ & 
    $0.828$ & 
    $\underline{\textbf{0.829}}$ & 
    $0.826$ \\
    HRV & 
    $0.700$ & 
    $0.713$ & 
    $\underline{\textbf{0.715}}$ & 
    $0.704$ \\
    KOR & 
    $0.516$ & 
    $\textbf{0.519}$ & 
    / &  
    / \\ 
    LAV & 
    $0.573$ & 
    $\textbf{0.588}$ & 
    $0.581$ & 
    $0.581$ \\
    RUS & 
    $0.752$ & 
    $0.753$ & 
    $\textbf{0.765}$ & 
    $0.760$ \\
    SLV & 
    $0.651$ & 
    $\underline{\textbf{0.696}}$ & 
    $0.695$ & 
    $0.669$ \\
    SWE & 
    $0.785$ & 
    $0.783$ & 
    $0.787$ & 
    $\textbf{0.793}$ \\
    ZHO & 
    $\textbf{0.787}$ & 
    $0.786$ &
    $0.784$ & 
    $0.781$ \\
	\bottomrule
	\end{tabular}
	\end{subtable}
    \begin{subtable}{0.47\textwidth}
    \centering
    \caption{BERT.}
    \begin{tabular}{@{\hskip 20pt} cccc}
    \toprule
         \makecell{+mBERT\\\vphantom{}\\\vphantom{}} & 
         \makecell{+mBERT\\+UPOS\\\vphantom{}} & 
         \makecell{+mBERT\\+UPOS\\+feats} & 
         \makecell{+mBERT\\+feats\\\vphantom{}} \\
    \midrule
    $\textbf{0.821}$ & 
    $0.816$ &
    $0.817$ & 
    $\textbf{0.821}$ \\
    $\textbf{0.948}$ & 
    $0.946$ & 
    $\textbf{0.948}$ & 
    $0.947$ \\
    $0.875$ & 
    $\textbf{0.879}$ & 
    $0.877$ & 
    $0.875$ \\
    $0.926$ & 
    $0.924$ & 
    $0.928$ & 
    $\textbf{0.929}$ \\
    $\textbf{0.874}$ & 
    $0.871$ & 
    $0.872$ & 
    $\textbf{0.874}$ \\
    $0.881$ & 
    $\textbf{0.882}$ &
    / &  
    / \\ 
    $0.766$ & 
    $0.776$ & 
    $\textbf{0.780}$ & 
    $0.777$ \\
    $0.868$ & 
    $0.868$ & 
    $\textbf{0.870}$ & 
    $0.869$ \\
    $0.848$ & 
    $0.849$ & 
    $0.851$ & 
    $\textbf{0.854}$ \\
    $0.885$ & 
    $\textbf{0.886}$ & 
    $0.885$ & 
    $0.880$ \\
    $\textbf{0.930}$ & 
    $\textbf{0.930}$ &
    $\textbf{0.930}$ & 
    $\textbf{0.930}$ \\
	\bottomrule
	\end{tabular}
	\end{subtable}
\end{table}

The baseline models involving BERT outperform their LSTM counterparts across all languages by a large margin.
When adding POS tags or universal features to the LSTM-based models, we observe an increase in performance over the baselines for nine languages. 
For five (Arabic, Korean, Latvian, Russian, and Swedish), the increase in the $F_1$ score is not statistically significant; for three of them, the difference is under 0.01. 
For the remaining four languages (Croatian, Estonian, Finnish, and Slovene), the increase is statistically significant and ranges from $0.015\%$ (Finnish and Croatian) to $0.045\%$ (Slovene). 
The two languages for which we observe no improvement after adding morphological information to the LSTM-based models are English and Chinese.

In BERT-based models, the additional information does not make a practical difference. 
For all languages except Slovene and Latvian, the increase in $F_1$ values over the baseline is below 0.005. However, we note that the better results on Slovene are possibly a result of the optimistically high quality of the additional morphological information. The Stanza model with which we obtained the predictions for Slovene was trained for a different task, but on the same dataset as we use for NER.
Still, the improvements are not statistically significant, so we do not explore the effect further.

\subsubsection{Results for the dependency parsing task}
\label{sec:dp-results}

For the DP evaluation task, we present LAS scores in Table \ref{tab:results-dp-las}. 
On the left, we show the results obtained with the parser containing additional LSTM embeddings, and on the right, with the parser containing additional BERT embeddings. 
For each, we report the results without and with additional POS tag and universal feature inputs (as introduced in \Cref{sec:depparse-models}).
We also statistically test the differences in the scores between the best performing enhanced variants and their baselines without morphological additions. As the splits are fixed in the DP tasks, we use the Z-test for the equality of two proportions~\citep{kanji2006-100statistical} at $p=0.01$ level.
The null hypothesis is that the scores of compared models are equal. We underline the best result for languages where the null hypothesis can be rejected.

\setlength{\tabcolsep}{3pt}
\begin{table}[htb]
    \caption{LAS scores for different models on the DP task in different languages using (a) LSTM or (b) BERT contextual embeddings. The best scores for each language and each type of contextual embeddings are marked with the bold typeface. Statistically significant differences at $p=0.01$ level in the best scores compared to the baseline of the same architecture are underlined.}
    \label{tab:results-dp-las}
    \begin{subtable}{0.47\textwidth}
    \centering
    \caption{LSTM.}
    \begin{tabular}{l @{\hskip 20pt} cccc}
    \toprule
         lang. & 
         \makecell{+LSTM\\\vphantom{}\\\vphantom{}} & 
         \makecell{+LSTM\\+UPOS\\\vphantom{}} & 
         \makecell{+LSTM\\+UPOS\\+feats} & 
         \makecell{+LSTM\\+feats\\\vphantom{}} \\
    \midrule
    ARA & 
    $76.74$ &                           
    $80.11$ &                           
    $\underline{\textbf{80.53}}$ &      
    $79.42$                             
    \\
    ENG & 
    $80.69$ &                           
    $\underline{\textbf{84.68}}$ &      
    $84.48$ &                           
    $82.96$ \\                          
    EST & 
    $76.24$ &                           
    $81.36$ &                           
    $\underline{\textbf{83.46}}$ &      
    $81.00$ \\                          
    FAS & 
    $84.42$ &                           
    $\underline{\textbf{87.59}}$ &      
    $87.32$ &                           
    $85.95$                             
    \\
    FIN & 
    $78.60$ &                           
    $83.06$ &                           
    $\underline{\textbf{83.97}}$ &      
    $81.16$ \\                          
    HEB & 
    $82.29$ &                           
    $\underline{\textbf{85.49}}$ &      
    $84.92$ &                           
    $83.84$                             
    \\
    HRV & 
    $79.59$ &                           
    $81.95$ &                           
    $\underline{\textbf{82.84}}$ &      
    $82.00$                             
    \\
    HUN  & 
    $69.58$ &                           
    $75.19$ &                           
    $\underline{\textbf{75.93}}$ &      
    $74.29$ \\                          
    KOR & 
    $74.46$ &                           
    $\underline{\textbf{82.35}}$ &      
    / &
    / \\
    LAV & 
    $78.08$ &                           
    $82.75$ &                           
    $\underline{\textbf{84.27}}$ &      
    $81.81$ \\                          
    LIT & 
    $63.65$ &                           
    $70.45$ &                           
    $\underline{\textbf{74.07}}$ &      
    $72.40$ \\                          
    RUS & 
    $79.24$ &                           
    $80.88$ &                           
    $\underline{\textbf{81.68}}$ &      
    $81.64$ \\                          
    SLV & 
    $83.03$ &                           
    $88.75$ &                           
    $\underline{\textbf{90.32}}$ &      
    $88.38$ \\                          
    SWE & 
    $78.97$ &                           
    $82.65$ &                           
    $\underline{\textbf{83.17}}$ &      
    $80.31$ \\                          
    TUR & 
    $66.85$ &                           
    $68.62$ &                           
    $\underline{\textbf{69.46}}$ &      
    $67.51$ \\                          
    ZHO & 
    $72.04$ &                           
    $78.65$ &                           
    $\underline{\textbf{78.89}}$ &      
    $71.09$ \\                          
    \bottomrule
    \end{tabular}
    \end{subtable}
    \begin{subtable}{0.47\textwidth}
    \centering
    \caption{BERT.}
    \begin{tabular}{@{\hskip 20pt} cccc}
    \toprule
         \makecell{+mBERT\\\vphantom{}\\\vphantom{}} & 
         \makecell{+mBERT\\+UPOS\\\vphantom{}} & 
         \makecell{+mBERT\\+UPOS\\+feats} & 
         \makecell{+mBERT\\+feats\\\vphantom{}} \\
    \midrule
    $80.97$ &                           
    $82.54$ &                           
    $\underline{\textbf{83.01}}$ &      
    $82.45$                             
    \\
    $88.09$ &                           
    $\underline{\textbf{89.35}}$ &      
    $89.13$ &                           
    $88.81$                             
    \\
    $84.30$ &                           
    $86.72$ &                           
    $\underline{\textbf{87.19}}$ &      
    $85.72$                             
    \\
    $89.06$ &                           
    $\underline{\textbf{90.80}}$ &      
    $90.65$ &                           
    $89.90$                             
    \\
    $86.07$ &                           
    $87.32$ &                           
    $\underline{\textbf{87.99}}$ &      
    $87.07$                             
    \\
    $87.18$ &                           
    $88.38$ &                           
    $\underline{\textbf{88.77}}$ &      
    $87.63$                             
    \\
    $86.37$ &                           
    $\underline{\textbf{87.63}}$  &     
    $87.44$ &                           
    $86.81$                             
    \\
    $77.07$ &                           
    $\underline{\textbf{80.04}}$ &      
    $79.96$ &                           
    $77.05$                             
    \\
    $85.46$ &                           
    $\underline{\textbf{87.59}}$ &      
    / &
    / 
    \\
    $83.02$ &                           
    $85.43$ &                           
    $\underline{\textbf{86.16}}$ &      
    $85.19$                             
    \\
    $72.62$ &                           
    $74.99$ &                           
    $\underline{\textbf{77.04}}$ &      
    $76.33$                             
    \\
    $85.99$ &                           
    $86.24$ &                           
    $\textbf{86.56}$ &                  
    $86.37$                             
    \\
    $91.67$ &                           
    $92.90$ &                           
    $\underline{\textbf{93.40}}$ &      
    $92.74$                             
    \\
    $86.67$ &                           
    $87.80$ &                           
    $\underline{\textbf{88.38}}$ &      
    $87.25$                             
    \\
    $70.60$ &                           
    $\underline{\textbf{72.99}}$ &      
    $72.58$ &                           
    $71.58$                             
    \\
    $79.49$ &                           
    $\underline{\textbf{83.94}}$ &      
    $83.21$ &                           
    $80.45$                             
    \\
    \bottomrule
    \end{tabular}
    \end{subtable}
\end{table}

Similarly to NER, the models involving BERT embeddings outperform the baselines involving LSTM embeddings on all $16$ languages by a large margin.
The models with the added POS tags or universal features noticeably improve over baselines with only LSTM embeddings for all languages.
The increase ranges between $2.44 \%$ (Russian) and $10.42 \%$ (Lithuanian). 
All compared differences between the LSTM baselines and the best enhanced variants are statistically significant at $p=0.01$. 

Contrary to the results observed on the NER task, adding morphological features to the models with BERT embeddings improves the performance scores for all languages. 
The increase ranges from $0.57 \%$ (Russian) and $4.45 \%$ (Chinese).
The increase is not statistically significant only for one language (Russian).
For the remaining $15$ languages, the increase is over one per cent and statistically significant.

Interestingly, for some languages, the enhanced parsers with LSTM come close in terms of LAS to the baseline parsers with BERT, For two languages (Lithuanian and Latvian), they even achieve better LAS scores.

\subsubsection{Results for the comment filtering task}
Table \ref{tab:comment-filtering-results} shows the CF results for LSTM and BERT baselines and their enhancements, described in Section~\ref{sec:comment-filtering-models}.
We compute the statistical significance of the differences between the baseline LSTM and BERT models and their best-performing counterparts with morphological additions. To do so, we use the unpaired t-test \citep{student1908-ttest} and check for statistical significance at $p=0.01$ level, as in the other two tasks.

\begin{table}[htb]
    \centering
    \caption{$F_1$ scores for baseline and enhanced models on the CF task in different languages. The left part of the table shows results for LSTM models, and the right part shows the results for BERT models. We report the mean and standard deviation over five runs. The highest mean score for each language is marked with the bold typeface separately for LSTM and BERT models. Note, however, that none of the improvements over the baselines is statistically significant at $p=0.01$ level.}
    \label{tab:comment-filtering-results}
    \begin{subtable}{0.47\textwidth}
    \caption{LSTM.}
    \begin{tabular}{l @{\hskip 15pt} cccc}
    \toprule
    lang. & 
    \makecell{+LSTM\\\vphantom{}\\\vphantom{}} & 
    \makecell{+LSTM\\+UPOS\\\vphantom{}} & 
    \makecell{+LSTM\\+UPOS\\+feats} & 
    \makecell{+LSTM\\+feats\\\vphantom{}} \\
    \midrule
    ARA &
    $\makecell{\textbf{0.752}\\(0.005)}$ & 
    $\makecell{0.748\\(0.006)}$ & 
    $\makecell{0.749\\(0.006)}$ & 
    $\makecell{0.749\\(0.006)}$ 
    \\[10pt]
    ELL &
    $\makecell{0.646\\(0.014)}$ & 
    $\makecell{0.647\\(0.012)}$ & 
    $\makecell{\textbf{0.663}\\(0.009)}$ & 
    $\makecell{0.652\\(0.014)}$ 
    \\[10pt]
    ENG & 
    $\makecell{0.881\\(0.012)}$ & 
    $\makecell{0.882\\(0.015)}$ & 
    $\makecell{0.869\\(0.012)}$ & 
    $\makecell{\textbf{0.883}\\(0.027)}$ \\[10pt] 
    KOR &
    $\makecell{\textbf{0.661}\\(0.012)}$ & 
    $\makecell{\textbf{0.661}\\(0.008)}$ & 
    / & 
    / \\[10pt]
    SLV &
    $\makecell{0.640\\(0.013)}$ & 
    $\makecell{\textbf{0.650}\\(0.015)}$ & 
    $\makecell{\textbf{0.650}\\(0.006)}$ & 
    $\makecell{0.649\\(0.007)}$ 
    \\[10pt]
    TUR &
    $\makecell{0.671\\(0.010)}$ & 
    $\makecell{\textbf{0.676}\\(0.010)}$ & 
    $\makecell{0.669\\(0.008)}$ & 
    $\makecell{\textbf{0.676}\\(0.009)}$ 
    \\[10pt]
    \bottomrule
    \end{tabular}
    \end{subtable}
    \begin{subtable}{0.47\textwidth}
    \caption{BERT.}
    \begin{tabular}{@{\hskip 15pt} cccc}
    \toprule
    \makecell{+mBERT\\\vphantom{}\\\vphantom{}} & 
    \makecell{+mBERT\\+UPOS\\\vphantom{}} & 
    \makecell{+mBERT\\+UPOS\\+feats} & 
    \makecell{+mBERT\\+feats\\\vphantom{}} \\
    \midrule
    $\makecell{\textbf{0.843}\\(0.005)}$ & 
    $\makecell{\textbf{0.843}\\(0.007)}$ & 
    $\makecell{0.840\\(0.008)}$ & 
    $\makecell{0.842\\(0.008)}$ 
    \\[10pt]
    $\makecell{0.820\\(0.022)}$ & 
    $\makecell{0.821\\(0.014)}$ & 
    $\makecell{\textbf{0.823}\\(0.015)}$ & 
    $\makecell{0.817\\(0.011)}$ 
    \\[10pt]
    $\makecell{0.922\\(0.012)}$ & 
    $\makecell{\textbf{0.923}\\(0.011)}$ & 
    $\makecell{\textbf{0.923}\\(0.020)}$ & 
    $\makecell{0.922\\(0.013)}$ \\[10pt] 
    $\makecell{0.722\\(0.007)}$ & 
    $\makecell{\textbf{0.732}\\(0.019)}$ & 
    / & 
    / \\[10pt]
    $\makecell{0.782\\(0.006)}$ & 
    $\makecell{0.784\\(0.005)}$ & 
    $\makecell{\textbf{0.785}\\(0.002)}$ & 
    $\makecell{0.782\\(0.005)}$ 
    \\[10pt]
    $\makecell{0.756\\(0.005)}$ & 
    $\makecell{0.762\\(0.008)}$ & 
    $\makecell{\textbf{0.765}\\(0.006)}$ & 
    $\makecell{0.762\\(0.010)}$ 
    \\[10pt]
    \bottomrule
    \end{tabular}
    \end{subtable}
\end{table}

All BERT-based models outperform the LSTM-based models by a large margin in all languages.
Adding POS tags and universal features to neural architectures of either type does not seem to significantly benefit their $F_1$ score in general.
Although the mean scores of the best performing enhanced models are often higher, the difference is under $0.010$ in most cases. The exceptions are Greek ($+0.017$) and Slovene ($+0.010$) for LSTM models, and Korean ($+0.010$) for BERT models, although the improvements are not statistically significant due to the noticeable standard deviation of the scores.
On the other hand, the enhanced models perform practically equivalently or slightly worse on Arabic.
As the performance scores do not statistically differ with the best set of hyperparameters, we do not further analyse the effect of different pooling types on the performance. However, we did not observe any pooling approach to perform best consistently.

\subsection{Additional experiments} 
To further analyse the impact of different aspects of the proposed morphological enhancements, we conducted several studies on the DP task, where the datasets and evaluation settings allow many experiments. Similarly as in \Cref{sec:dp-results}, we statistically evaluate differences in performance between the baseline model and the best enhancement using the Z-test for the equality of two proportions. In cases where the null hypothesis can be rejected at $p=0.01$ level, we underline the respective compared score. We test the impact of additional training time (\Cref{sec:time}), quality of morphological information (\Cref{sec:quality}), and different variants of BERT models (\Cref{sec:BERTvariants}). 

\subsubsection{Additional training time} \label{sec:time}
To test if the observed differences in performance are due to random variation in the training of models, which could be reduced with longer training times, we increase the maximum training time from $10$ to $15$ epochs. 
We show the results in Table \ref{tab:results-dp-longer-training}.

\setlength{\tabcolsep}{3pt}
\begin{table}[htb]
    \caption{LAS scores achieved by models that are trained for up to $5$ additional epochs (a maximum training time of $15$ epochs instead of $10$). The results for $10$ epochs are presented in Table \ref{tab:results-dp-las}. Statistically significant differences in best scores at $p=0.01$ level compared to the baselines of the same architecture are underlined.}
    \label{tab:results-dp-longer-training}
    \begin{subtable}{0.47\textwidth}
    \centering
    \caption{LSTM.}
    \begin{tabular}{l @{\hskip 20pt} cccc}
    \toprule
         lang. & 
         \makecell{+LSTM\\\vphantom{}\\\vphantom{}} & 
         \makecell{+LSTM\\+UPOS\\\vphantom{}} & 
         \makecell{+LSTM\\+UPOS\\+feats} & 
         \makecell{+LSTM\\+feats\\\vphantom{}} \\
    \midrule
    ARA & 
    $78.00$ &                           
    $80.56$ &                           
    $\underline{\textbf{81.13}}$ &      
    $79.68$                             
    \\
    ENG & 
    $81.47$ &                           
    $84.74$ &                           
    $\underline{\textbf{85.26}}$ &      
    $83.58$ \\                          
    EST & 
    $76.72$ &                           
    $81.52$ &                           
    $\underline{\textbf{83.65}}$ &      
    $81.24$ \\                          
    FAS & 
    $84.82$ &                           
    $\underline{\textbf{87.61}}$ &      
    $87.32$ &                           
    $85.59$ \\                          
    FIN & 
    $80.28$ &                           
    $83.77$ &                           
    $\underline{\textbf{84.22}}$ &      
    $81.95$ \\                          
    HEB & 
    $83.06$ &                           
    $\underline{\textbf{86.35}}$ &      
    $85.95$ &                           
    $84.83$ \\                          
    HRV & 
    $79.49$ &                           
    $82.72$ &                           
    $\underline{\textbf{83.26}}$ &      
    $82.74$                             
    \\
    HUN  & 
    $70.41$ &                           
    $76.18$ &                           
    $\underline{\textbf{77.43}}$ &      
    $75.59$ \\                          
    KOR & 
    $74.41$ &                           
    $\underline{\textbf{82.28}}$ &      
    / &
    / \\
    LAV & 
    $79.58$ &                           
    $83.32$ &                           
    $\underline{\textbf{84.71}}$ &      
    $82.50$ \\                          
    LIT & 
    $66.48$ &                           
    $71.03$ &                           
    $74.22$ &                           
    $\underline{\textbf{74.28}}$ \\     
    RUS & 
    $79.79$ &                           
    $81.35$ &                           
    $82.18$ &                           
    $\underline{\textbf{82.21}}$ \\     
    SLV & 
    $84.72$ &                           
    $89.19$ &                           
    $\underline{\textbf{90.64}}$ &      
    $88.60$ \\                          
    SWE & 
    $79.97$ &                           
    $84.58$ &                           
    $\underline{\textbf{85.23}}$ &      
    $82.10$ \\                          
    TUR & 
    $67.68$ &                           
    $69.86$ &                           
    $\underline{\textbf{70.42}}$ &      
    $69.12$ \\                          
    ZHO & 
    $72.94$ &                           
    $\underline{\textbf{80.45}}$ &      
    $79.60$ &                           
    $72.69$ \\                          
    \bottomrule
    \end{tabular}
    \end{subtable}
    \begin{subtable}{0.47\textwidth}
    \centering
    \caption{BERT.}
    \begin{tabular}{@{\hskip 20pt} cccc}
    \toprule
         \makecell{+mBERT\\\vphantom{}\\\vphantom{}} & 
         \makecell{+mBERT\\+UPOS\\\vphantom{}} & 
         \makecell{+mBERT\\+UPOS\\+feats} & 
         \makecell{+mBERT\\+feats\\\vphantom{}} \\
    \midrule
    $81.75$ &                           
    $82.77$ &                           
    $\underline{\textbf{83.43}}$ &      
    $82.95$                             
    \\
    $87.77$ &                           
    $\underline{\textbf{89.56}}$ &      
    $89.49$ &                           
    $88.89$                             
    \\
    $84.60$ &                           
    $86.60$ &                           
    $\underline{\textbf{87.26}}$ &      
    $85.78$                             
    \\
    $88.90$ &                           
    $\underline{\textbf{90.88}}$ &      
    $90.75$ &                           
    $89.94$                             
    \\
    $87.06$ &                           
    $87.98$ &                           
    $\underline{\textbf{88.42}}$ &      
    $87.85$                             
    \\
    $88.02$ &                           
    $88.88$ &                           
    $\underline{\textbf{89.34}}$ &      
    $88.83$                             
    \\
    $87.03$ &                           
    $87.55$ &                           
    $\textbf{87.71}$ &                  
    $86.97$                             
    \\
    $79.31$ &                           
    $81.29$ &                           
    $\underline{\textbf{81.88}}$ &      
    $79.71$                             
    \\
    $85.58$ &                           
    $\underline{\textbf{87.84}}$ &      
    / &
    / 
    \\
    $84.04$ &                           
    $85.66$ &                           
    $\underline{\textbf{86.73}}$ &      
    $85.19$                             
    \\
    $74.17$ &                           
    $76.11$ &                           
    $\underline{\textbf{77.88}}$ &      
    $77.04$                             
    \\
    $86.15$ &                           
    $87.04$ &                           
    $\textbf{87.23}$ &                  
    $86.64$                             
    \\
    $92.11$ &                           
    $92.80$ &                           
    $\underline{\textbf{93.78}}$ &      
    $93.08$                             
    \\
    $87.43$ &                           
    $\underline{\textbf{88.86}}$ &      
    $88.77$ &                           
    $87.72$                             
    \\
    $71.78$ &                           
    $71.82$ &                           
    $\underline{\textbf{73.38}}$ &      
    $72.24$                             
    \\
    $81.29$ &                           
    $\underline{\textbf{84.19}}$ &      
    $84.01$ &                           
    $80.85$                             
    \\
    \bottomrule
    \end{tabular}
    \end{subtable}
\end{table}

We can observe that longer training times slightly increases the scores for all model variants, though their relative order stays the same. The models with added morphological features still achieve better results, so the performance increases do not seem to be the effect of random fluctuations in training due to the number of training steps.
All improvements over baselines of the parsers using LSTM embeddings remain statistically significant. In contrast, for the parsers using BERT embeddings, the improvements for two languages are now no longer statistically significant (Croatian and Russian).

\subsubsection{Quality of morphological information} \label{sec:quality}
In the second additional experiment, we evaluate the impact of the quality of morphological information. We replace the high-quality (human-annotated) POS tags and morphological features used in \Cref{sec:depparse-models} with those predicted by machine learning models. In this way, we test a realistic setting where the morphological information is at least to a certain degree noisy.
We obtain POS tags and morphological features from Stanza models prepared for the tested languages \citep{qi2020stanza}. To avoid overly optimistic results, we use models that are not trained on the same datasets used in our DP experiments. This is possible for a subset of nine languages. We note the used Stanza models in Appendix \ref{appendix:stanza-models}, together with the proportion of tokens, for which POS tags and \emph{all} universal features are correctly predicted (i.e. their accuracy).

We show the results of DP models, trained with predicted morphological features in Table \ref{tab:dp-predicted-features}.

\setlength{\tabcolsep}{3pt}
\begin{table}[htb]
    \caption{LAS scores achieved by DP models that are trained with predicted (noisy) instead of human-annotated morphological features. We provide the accuracy of predicted UPOS tags and universal features in Appendix \ref{appendix:stanza-models}. Statistically significant differences in the best scores compared to baselines at $p=0.01$ level are underlined.}
    \label{tab:dp-predicted-features}
    \begin{subtable}{0.47\textwidth}
    \centering
    \caption{LSTM.}
    \begin{tabular}{l @{\hskip 20pt} cccc}
    \toprule
         lang. & 
         \makecell{+LSTM\\\vphantom{}\\\vphantom{}} & 
         \makecell{+LSTM\\+UPOS\\\vphantom{}} & 
         \makecell{+LSTM\\+UPOS\\+feats} & 
         \makecell{+LSTM\\+feats\\\vphantom{}} \\
    \midrule
    ENG & 
    $80.69$ & 
    $\textbf{81.36}$ & 
    $81.02$ & 
    $81.22$ \\
    EST & 
    $76.24$ &
    $77.15$ & 
    $\underline{\textbf{77.39}}$ & 
    $76.19$ \\
    FIN & 
    $78.60$ & 
    $80.61$ & 
    $\underline{\textbf{81.01}}$ & 
    $80.43$ \\
    KOR & 
    $74.46$ & 
    $\underline{\textbf{76.23}}$ & 
    / & 
    / \\
    FAS & 
    $84.42$ & 
    $\textbf{85.04}$ & 
    $84.07$ & 
    $84.59$ \\
    RUS &
    $79.24$ & 
    $79.53$ & 
    $\underline{\textbf{80.89}}$ & 
    $79.56$ \\
    SLV & 
    $83.03$ &
    $84.06$ & 
    $\underline{\textbf{85.00}}$ & 
    $84.12$ \\
    SWE & 
    $78.97$ & 
    $80.06$ & 
    $\underline{\textbf{80.08}}$ & 
    $79.24$ \\
    TUR & 
    $66.85$ & 
    $67.42$ & 
    $\textbf{68.03}$ & 
    $67.76$ \\
    \bottomrule
    \end{tabular}
    \end{subtable}
    \begin{subtable}{0.47\textwidth}
    \centering
    \caption{BERT.}
    \begin{tabular}{@{\hskip 20pt} cccc}
    \toprule
     \makecell{+mBERT\\\vphantom{}\\\vphantom{}} & 
     \makecell{+mBERT\\+UPOS\\\vphantom{}} & 
     \makecell{+mBERT\\+UPOS\\+feats} & 
     \makecell{+mBERT\\+feats\\\vphantom{}} \\
    \midrule
    $88.09$ & 
    $\textbf{88.33}$ & 
    $88.05$ & 
    $87.74$ \\
    $84.30$ & 
    $\textbf{84.69}$ & 
    $84.39$ & 
    $84.42$ \\
    $86.07$ & 
    $86.33$ & 
    $\textbf{86.73}$ & 
    $85.91$ \\
    $85.46$ & 
    $\textbf{85.90}$ & 
    / & 
    / \\
    $89.06$ & 
    $\textbf{89.28}$ & 
    $88.82$ & 
    $88.84$ \\
    $85.99$ & 
    $\textbf{86.17}$ & 
    $85.74$ & 
    $85.27$ \\
    $\textbf{91.67}$ & 
    $91.42$ & 
    $91.57$ & 
    $91.23$  \\
    $86.67$ & 
    $86.59$ & 
    $\textbf{86.96}$ & 
    $86.33$ \\
    $70.60$ & 
    $\textbf{71.26}$ & 
    $71.11$ & 
    $70.77$ \\
    \bottomrule
    \end{tabular}
    \end{subtable}
\end{table}

The general trend is that using predicted features results in much smaller (best case) performance increases, though some languages still see significant increases.
For LSTM models, the increases range from $0.62\%$ (Persian) to $2.41\%$ (Finnish), with six out of nine being statistically significant.
For BERT models, the increases are all statistically insignificant and range from $-0.10\%$ (i.e. decrease, Slovene) to $0.66\%$ (Finnish).
These results are consistent with the results of our NER experiments in \Cref{sec:ner-results}, where we have no access to human-annotated features and find that noisy features only help LSTM-based models. 


These results indicate that adding predicted morphological features to models with BERT embeddings might not be practically useful, since their quality needs to be very high.
However, since human-annotated morphological features improve the performance on the DP task, this suggests that there could be room for improvement in BERT pre-training. It seems that the pre-training tasks of BERT (masked language modelling and next sentence prediction) do not fully capture the morphological information present in the language. However, it is unlikely that the models could capture all information present in the ground truth annotations, as humans can disambiguate the grammatical role of a word even where syncretism occurs.

\subsubsection{Variants of BERT model} \label{sec:BERTvariants}
In the third additional experiment, we revert to using the ground truth morphological annotations but replace the embeddings obtained from the multilingual uncased BERT model with those obtained from more specific multilingual BERT models and monolingual BERT models.
In experiments involving multilingual BERT models, we test the Croatian/Slovene/English CroSloEngual BERT, Finnish/Estonian/English FinEst BERT \citep{ulcar2020finest}, and Bulgarian/Czech/Polish/Russian Slavic BERT \citep{arkhipov-2019-slavicbert}. 
In experiments with monolingual BERT models, we use the Arabic bert-base-arabic \citep{safaya-2020-arabertbase}, English bert-base-cased \citep{devlin2019-bert}, Estonian ESTBert \citep{tanvir-2021-estbert}, Finnish FinBERT \citep{virtanen2019multilingual}, Hebrew AlephBERT \citep{seker-2021-alephbert}, Hungarian huBERT \citep{Nemeskey-2021a-hubert}, Korean bert-kor-base, Persian ParsBERT \citep{farahani-2020-parsbert}, Russian RuBert \citep{kuratov2019-rubert}, Swedish bert-base-swedish-cased, Turkish BERTurk, and Chinese bert-base-chinese models. 

We only performed the experiments for a subset of studied languages for which we were able to find more specific BERT models.
The aim is to check if the additional morphological features improve the performance of less general, i.e. more language-specific BERT models. 
These are trained on a lower number of languages, and larger amounts of texts in the included languages, compared to the original multilingual BERT model \citep{devlin2019-bert} that was trained on 104 languages simultaneously. Due to this language-specific training, we expect these BERT models to capture the nuances of the languages better, thus possibly benefiting less from the additional morphological features. 

We show the results in Table \ref{tab:dp-more-specific-bert}.
In most cases, the specific multilingual BERT models, even without additional features, do as well as or better than the best performing original multilingual BERT model with additional features. The only worse LAS scores are achieved on Russian.
This indicates that the more specific multilingual BERT models are generally better suited for the DP task than the original multilingual models.
The addition of morphological features increases the LAS even further. 
The improvements range from $0.54\%$ (Slovene) to $2.23\%$ (Estonian) and are statistically significant for four out of six languages.

\setlength{\tabcolsep}{3pt}
\begin{table}[htb]
    \centering
    \caption{LAS scores in the DP task achieved by more specific multilingual (top), and monolingual (bottom) BERT models. The more specific multilingual BERT models were trained on a smaller set of languages (three or five) than the original multilingual BERT model ($104$). For the base parsers with BERT embeddings, we display the improvement in LAS over using the original multilingual BERT for $104$ languages ($\uparrow_{mtl}$). Statistically significant differences of best scores to mBERT baselines at $p=0.01$ level are underlined.}
    \label{tab:dp-more-specific-bert}
    \begin{tabular}{ll rl@{\hskip 15px} ccc}
    \toprule
    lang. &
    model handle &
    \makecell{+mBERT\\\vphantom{}\\\vphantom{}} & 
    \makecell{($\uparrow_{mtl}$)\\\vphantom{}\\\vphantom{}} &
    \makecell{+mBERT\\+UPOS\\\vphantom{}} & 
    \makecell{+mBERT\\+UPOS\\+feats} & 
    \makecell{+mBERT\\+feats\\\vphantom{}} \\
    \bottomrule
    HRV & CroSloEngual BERT & 
    $87.84$ & ($+1.47$) &
    $\underline{\textbf{89.01}}$ &
    $88.33$ & 
    $87.98$ \\
    
    ENG & CroSloEngual BERT & 
    $88.37$ & ($+0.28$) &
    $\underline{\textbf{89.69}}$ & 
    $89.65$ & 
    $88.58$ \\
    
    SLV & CroSloEngual BERT &
    $93.98$ & ($+2.31$) &
    $94.27$ & 
    $\textbf{94.52}$ & 
    $94.05$ \\
    
    FIN & FinEst BERT & 
    $89.35$ & ($+3.28$) &
    $\underline{\textbf{90.66}}$ & 
    $90.61$ & 
    $90.54$ \\
    
    EST & FinEst BERT &
    $86.51$ & ($+2.21$) &
    $88.67$ & 
    $\underline{\textbf{88.74}}$ & 
    $87.44$ \\
    
    RUS & Slavic BERT &
    $85.60$ & ($-0.39$) &
    $86.32$ & 
    $\textbf{86.39}$ & 
    $86.39$ \\
    \midrule
    ARA & arabic-bert-base & 
    $83.40$ & ($+2.43$) &
    $84.08$ & 
    $\underline{\textbf{84.93}}$ & 
    $84.15$ \\
    
    ENG & bert-base-cased & 
    $88.61$ & ($+0.52$) &
    $\underline{\textbf{89.69}}$ & 
    $89.56$ & 
    $88.59$ \\
    
    EST & EstBERT &
    $86.67$ & ($+2.37$) &
    $88.69$ & 
    $\underline{\textbf{88.80}}$ & 
    $87.32$ \\
    
    FIN & FinBERT & 
    $91.51$ & ($+5.44$) &
    $\textbf{92.19}$ & 
    $91.73$ & 
    $91.80$ \\
    
    HEB & AlephBERT  &
    $89.41$ & ($+2.23$) &
    $\textbf{90.37}$ & 
    $89.31$ & 
    $89.71$ \\
    
    HUN & huBERT &
    $81.28$ & ($+4.21$) &
    $\underline{\textbf{82.88}}$ & 
    $81.93$ & 
    $81.83$ \\
    
    KOR & bert-kor-base &
    $88.51$ & ($+3.05$) &
    $\underline{\textbf{89.92}}$ & 
    / & 
    / \\
    
    FAS & ParsBERT & 
    $91.10$ & ($+2.04$) &
    $\underline{\textbf{92.22}}$ & 
    $92.20$ & 
    $91.69$ \\
    
    RUS & RuBert &
    $86.39$ & ($+0.40$) &
    $\textbf{87.31}$ & 
    $87.04$ & 
    $86.61$ \\
    
    SWE & bert-base-swedish-cased &
    $89.41$ & ($+2.74$) &
    $\underline{\textbf{90.63}}$ & 
    $90.53$ & 
    $90.21$ \\
    
    TUR & BERTurk &
    $76.78$ & ($+6.18$) &
    $76.75$ & 
    $\textbf{76.87}$ & 
    $76.07$ \\
    
    ZHO & bert-base-chinese &
    $84.72$ & ($+5.23$) &
    $\underline{\textbf{86.31}}$ & 
    $85.81$ & 
    $84.17$ \\
    \bottomrule
    \end{tabular}
\end{table}

The monolingual models without additional features set an even higher baseline performance. 
The results of including additional information are mixed, though surprisingly many languages still see a significant increase in LAS.
Out of twelve languages, the improvements are significant for eight and not significant for four languages.

These results indicate that the additional morphological features contain valuable information for the DP task, which the more specific BERT models still do not capture entirely. However, they improve over massively multilingual (i.e. more general) models. 
We suspect the increase would be even less pronounced if we experimented with ``large`` (as opposed to base-sized) variants, e.g., the large English BERT would likely benefit even less from additional morphological information. We leave this line of experiments for further work. 

\section{Conclusion}
\label{sec:conclusion}

We analysed adding explicit morphological information in the form of embeddings for POS tags and universal features to two currently dominant neural network architectures used in NLP: LSTM networks and transformer-based BERT models. We compared models enhanced with morphological information with baselines on three tasks (NER, DP, and CF). To obtain general conclusions, we used a variety of morphologically-rich languages from different language families. We make the code to re-run our experiments publicly available\footnote{\url{https://github.com/matejklemen/morphological-dependency-parsing} (DP),\\
\url{https://github.com/EMBEDDIA/morphological-fasttext} (NER),\\
\url{https://github.com/EMBEDDIA/morphological-BERT} (NER),\\
 \url{https://github.com/matejklemen/morphological-comment-filtering} (CF)}.

The results indicate that adding morphological information to CF prediction models is not beneficial, but it improves the performance in the NER and DP tasks. For the DP task, the improvement depends on the quality of the morphological features. The additional morphological features consistently benefited LSTM-based models for NER and DP, both when they were of high quality and predicted (noisy).
For BERT-based models, the \emph{predicted} features do not make any practical difference for the NER and DP task but improve the performance in the DP task when they are of \emph{high quality}. Testing different variants of BERT shows that language-specialised variants enhance the performance on the DP task and the additional morphological information is still beneficial, although less and less as we shift from multilingual towards monolingual models.

Comparing different BERT variants indicates that BERT models do not entirely capture the language morphology. Since the release of BERT, several new pre-training objectives have been proposed, such as syntactic and semantic phrase masking~\citep{zhou2020limitbert} and span masking~\citep{joshi-spanbert-2020}.
In further work, it makes sense to apply these models to the DP task to test how well they capture the morphology. Further, the effect of morphological features could be analysed on additional tasks and languages since the explicit morphological information does not seem to benefit them equally.

 \subsection*{Acknowledgements}
This work was supported by European Union’s Horizon 2020 Programme project EMBEDDIA (Cross-Lingual Embeddings for Less-Represented Languages in European News Media, grant no. 825153). The research was supported by the Slovene Research Agency through research core funding no. P6-0411 and the young researcher grant as well the Ministry of Culture of the Republic of Slovenia through project Development of Slovene in Digital Environment (RSDO). The Titan X Pascal used for a part of this research was donated by the NVIDIA Corporation.

\bibliographystyle{plainnat}  
\bibliography{references}

\clearpage
\appendix
\section{Used Stanza models}
\label{appendix:stanza-models}

\setlength{\tabcolsep}{3pt}
\begin{table}[htb]
    \small
    \centering
    \begin{subtable}{0.25\textwidth}
    \centering
    \caption{Stanza models used in the NER experiments.}
    \label{tab:stanza-models-ner}
    \begin{tabular}{lc}
    \toprule
    \makecell{lang.\\\vphantom{}} & \makecell{Model\\\vphantom{}} \\
    \midrule
    Arabic & padt \\
    Chinese & gsdsimp \\
    Croatian & set \\
    English & ewt \\
    Estonian & edt \\
    Finnish & tdt \\
    Korean & kaist \\
    Latvian & lvtb \\
    Russian & syntagrus \\
    Slovene & ssj \\
    Swedish & talbanken \\
    \bottomrule
    \end{tabular}
    \end{subtable}
    \begin{subtable}{0.4\textwidth}
    \centering
    \caption{Stanza models used in the additional DP experiments.}
    \label{tab:stanza-models-dp}
    \begin{tabular}{lccc}
    \toprule
    lang. & Model & \makecell{accuracy\\(UPOS)} & \makecell{accuracy\\(UFeats)} \\
    \midrule
    English & gum & $92.45$ & $93.92$ \\
    Estonian & ewt & $91.15$ & $88.86$ \\
    Finnish & ftb & $87.59$ & $86.20$ \\
    Korean & gsd & $70.81$ & / \\
    Persian & seraji & $82.32$ & $76.30$ \\
    Russian & syntagrus & $89.32$ & $85.22$ \\
    Slovene & sst & $80.45$ & $78.74$ \\
    Swedish & lines & $94.66$ & $86.16$ \\
    Turkish & imst & $86.20$ & $68.83$ \\
    \bottomrule
     & \\
     & \\
    \end{tabular}
    \end{subtable}
    \begin{subtable}{0.25\textwidth}
    \centering
    \caption{Stanza models used in the CF experiments.}
    \label{tab:stanza-models-cf}
    \begin{tabular}{lc}
    \toprule
    \makecell{lang.\\\vphantom{}} & \makecell{Model\\\vphantom{}} \\
    \midrule
    Arabic & padt \\
    English & ewt \\
    Greek & gdt \\
    Korean & gsd \\
    Slovene & ssj \\
    Turkish & imst \\
    \bottomrule
    & \\
    & \\
    & \\
    & \\
    & \\
    \end{tabular}
    \end{subtable}
\end{table}

\section{Results of dependency parsing with cased multilingual BERT}
\label{appendix:cased-mtl-bert}

Table \ref{tab:dp-cased-mtl-bert} shows the results of a subset of the main dependency parsing experiments performed using a \textbf{cased} multilingual BERT model (bert-base-multilingual-cased). However, the conclusions drawn from \Cref{tab:results-dp-las} and  \Cref{tab:dp-cased-mtl-bert} are the same. 

\setlength{\tabcolsep}{3pt}
\begin{table}[htb]
    \centering
    \caption{LAS achieved using a cased (instead of uncased) multilingual BERT model on a random sample of eight languages in the DP task. Statistically significant differences between baselines and best scores at $p=0.01$ level are underlined.}
    \label{tab:dp-cased-mtl-bert}
    \begin{tabular}{l cccc}
    \toprule
    lang. &
    \makecell{+mBERT\\\vphantom{}\\\vphantom{}} & 
    \makecell{+mBERT\\+UPOS\\\vphantom{}} & 
    \makecell{+mBERT\\+UPOS\\+feats} & 
    \makecell{+mBERT\\+feats\\\vphantom{}} \\
    \midrule
    ZHO & 
    $83.03$ & 
    $\underline{\textbf{84.97}}$ & 
    $84.23$ &
    $82.53$ \\
    ENG & 
    $88.04$ & 
    $\textbf{88.74}$ & 
    $88.59$ &
    $88.19$ \\
    FIN & 
    $85.65$ & 
    $\underline{\textbf{87.55}}$ & 
    $87.34$ &
    $87.17$ \\
    HUN & 
    $77.77$ & 
    $80.14$ & 
    $\underline{\textbf{80.40}}$ &
    $77.80$ \\
    FAS & 
    $88.99$ & 
    $\underline{\textbf{90.66}}$ & 
    $90.57$ &
    $89.96$ \\
    SLV & 
    $91.35$ & 
    $92.46$ & 
    $\underline{\textbf{93.16}}$ &
    $92.09$ \\
    SWE & 
    $85.66$ & 
    $\underline{\textbf{87.23}}$ & 
    $87.16$ &
    $86.30$ \\
    TUR & 
    $71.52$ & 
    $\underline{\textbf{72.82}}$ & 
    $72.38$ &
    $70.92$ \\
    \bottomrule
    \end{tabular}
\end{table}

\label{lastpage}

\end{document}